\documentclass[runningheads]{llncs}

\usepackage{eccv}

\usepackage{eccvabbrv}

\usepackage{graphicx}
\usepackage{booktabs}

\usepackage[accsupp]{axessibility}  

\usepackage{hyperref}

\usepackage{orcidlink}

\usepackage{comment}
\usepackage{pifont}
\usepackage{colortbl}
\usepackage{xcolor} 
\usepackage{mdframed}

\definecolor{headergray}{gray}{0.9}
\definecolor{sectiongray}{gray}{0.95}
\definecolor{ourrow}{RGB}{230,245,255}

\begin{document}

\title{LongEgoRefer: A Benchmark for Long-Form Egocentric Video Referring Expression Comprehension} 

\titlerunning{LongEgoRefer}

\author{Shunya Kato\inst{1}\thanks{Woven by Toyota was not involved in this work. The views and opinions expressed in this paper are those of the authors and do not necessarily reflect the official policy or position of Woven by Toyota.}\orcidlink{0009-0008-8603-1751} \and
Taiki Miyanishi\inst{2}\orcidlink{0000-0001-9105-1601} \and
Shuhei Kurita\inst{3,4,5}\orcidlink{0000-0001-7415-3120} \and
Mahiro Ukai\inst{4}\orcidlink{0009-0001-6266-9477} \and
Nakamasa Inoue\inst{4}\orcidlink{0000-0002-9761-4142} \and
Chenhui Chu\inst{5,6}\orcidlink{0000-0001-9848-6384}
}

\authorrunning{S. Kato et al.}

\institute{Woven by Toyota, Japan \and The University of Tokyo, Japan \and National Institute of Informatics, Japan \and Institute of Science Tokyo, Japan \and NII LLMC, Japan \and Kyoto University, Japan}

\maketitle

\vspace{-3mm}

\begin{abstract}

Egocentric videos capture rich and diverse human–object interactions and have emerged as a fundamental resource for understanding human activities related to objects. In this context, Video Referring Expression Comprehension (Video REC), the task of localizing the temporal and spatial extent of a referred object in video frames given a natural language query, plays a key role in linking textual descriptions to observed objects in untrimmed egocentric recordings. However, existing egocentric Video REC benchmarks primarily focus on short video clips, where some target object appears densely within frames. Such settings do not reflect real-world egocentric recordings, which are long-form, untrimmed, and characterized by sparse object occurrences and complex activity transitions.
To address this limitation, we introduce LongEgoRefer, a novel and challenging benchmark constructed from long-form videos in the Ego4D dataset. LongEgoRefer contains 1,498 referring expressions with an average video duration of 45 minutes. The benchmark exhibits extreme target sparsity, detailed linguistic descriptions, and complex human–object interactions embedded in long, dynamic egocentric narratives. Consequently, it defines a demanding spatio-temporal grounding problem that requires models to identify both when an event occurs and where the referred object appears within extended video sequences.
We evaluate existing Video REC approaches, including training-free baselines based on vision–language models combined with Grounded SAM2. Extensive experiments show that even advanced baselines and current state-of-the-art models struggle significantly on LongEgoRefer. These results highlight the intrinsic difficulty of long-form egocentric spatio-temporal grounding and emphasize the need for more robust video understanding models. Our benchmark and code are available at~\url{https://github.com/shunya-kato/LongEgoRefer}.

\end{abstract}    
\section{Introduction}
\label{sec:intro}

\begin{figure*}[t]
\centering
\includegraphics[width=\linewidth]{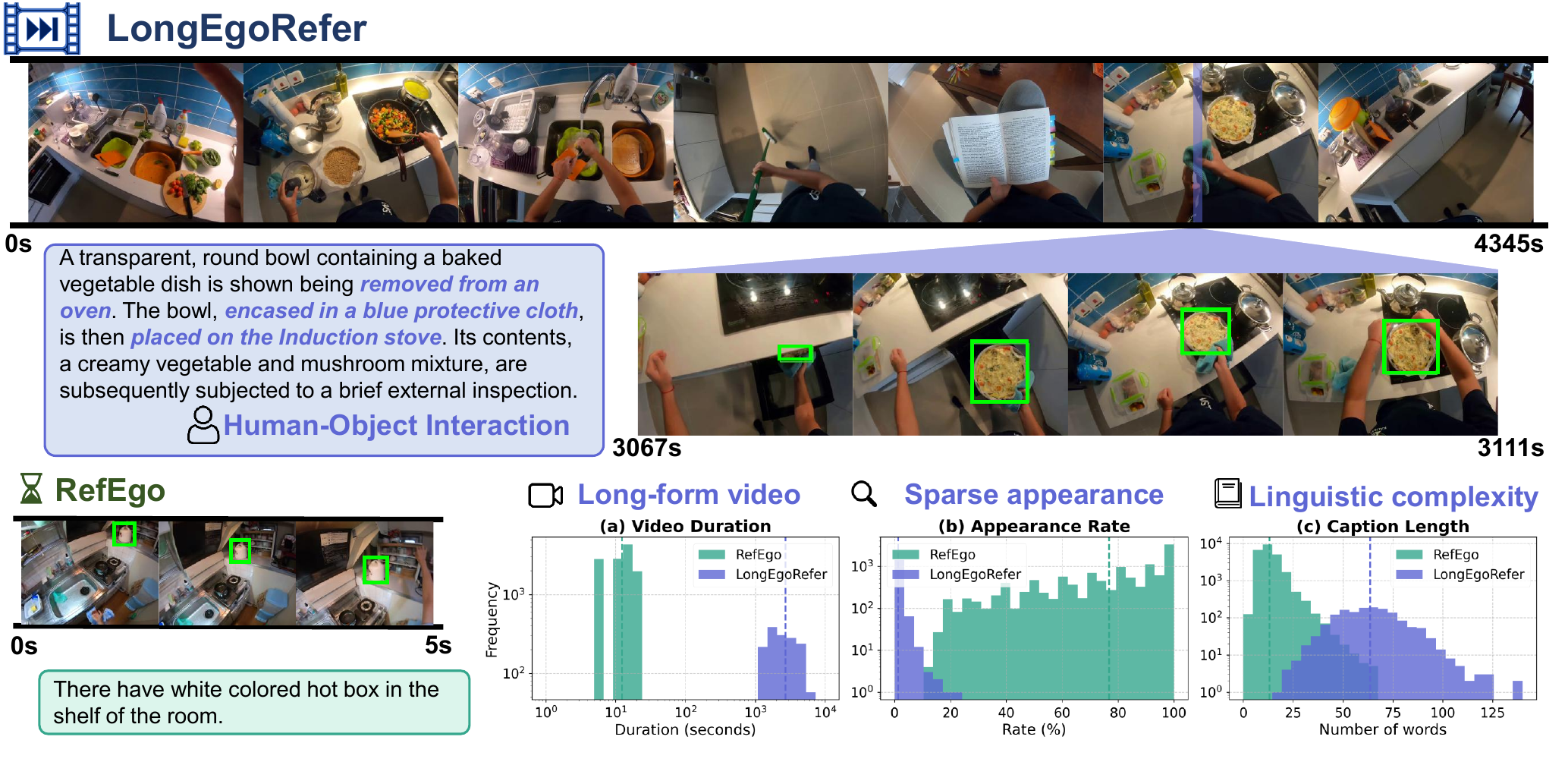}
\caption{Data comparison of LongEgoRefer and RefEgo. (a) Video durations in LongEgoRefer are orders of magnitude longer than those in RefEgo. (b) The appearance rate is significantly lower and sparser in LongEgoRefer compared to RefEgo. (c) To handle the complexity of long-form videos, captions in LongEgoRefer are substantially longer and more descriptive than those in RefEgo. Dashed lines indicate the mean value for each distribution.}
\label{fig:refego_longegorefer_comparison}
\end{figure*}

Video Referring Expression Comprehension (Video REC) is a critical task in computer vision that involves localizing a specific object or region within a video based on a natural language query, known as \textit{referring expression}. The objective of this task is to perform spatio-temporal localization of the target object indicated by a textual query such as ``the blue chair spinning on the left side of the screen'' across all frames of a given video. This is a fundamental and challenging multimodal problem that requires not only recognizing an object but also understanding its attributes, spatial relationships, and dynamic context, including temporal changes and movements, thereby ``grounding'' language to visual information. While significant progress has been made in REC for static images, extending this task to videos introduces new dimensions of complexity. It demands that a model possess advanced spatio-temporal reasoning capabilities to capture appearance changes of the object over time and the dynamics of complex interactions between multiple objects.

Among the various forms of video, the importance of the egocentric recordings of the first-person views has grown rapidly in recent years~\cite{Damen_2018_ECCV, grauman2022ego4d,grauman2024ego,Perrett2025_HD-EPIC}, driven by the anticipated proliferation of AR/VR devices, smart glasses~\cite{engel2023project}, and collaborative robots~\cite{bjorck2025gr00t}. In contrast to third-person videos that provide observers' viewpoint, egocentric videos provide the camera wearers' subjective experience. The camera motion directly reflects the wearer's focus of attention, while the appearance of their hands in the frame captures their physical interactions with the environment, providing an unparalleled source of information for understanding user intent.
As such, REC in the egocentric view is a foundational technology for the seamless integration of human-centric AI systems into our daily lives and professional workspaces~\cite{kurita2023refego,Tang_NeurIPS23_EgoTracks}. However, research in this domain is still in its nascent stages, with a notable lack of benchmarks that reflect the complexity of the real world. The unique and challenging characteristics of the egocentric perspective—such as drastic viewpoint changes, frequent occlusions by the wearer's hands, and long-duration tasks—pose significant challenges to existing methods.

While previous research has made significant progress, existing benchmarks do not adequately capture the complex situations that humans encounter in the real world. Specifically, there are three main limitations or gaps:
1. Lack of Temporal Context: Many existing benchmarks are composed of short video clips, typically lasting from a few to tens of seconds. However, real-world tasks often require long-form memory and contextual understanding.
2. Simplified Scenarios: The tasks tend to be oversimplified, with the target object often being clearly visible for the majority of the video. In the real world, it is common for an object to appear only transiently or be situated among many confusing objects (distractors).
3. Neglect of Interaction: Most importantly, there is a significant lack of referring expressions that focus on dynamic and complex human-object interactions, such as a person manipulating an object or collaborating with others. This capability is essential for the next generation of AI assistants to be truly useful.

To overcome these challenges, in this work, we propose LongEgoRefer, a new egocentric video REC benchmark based on more realistic and challenging scenarios (see Figure~\ref{fig:refego_longegorefer_comparison}). This benchmark was created by combining long-form videos from Ego4D with the precise object tracking annotations from EgoTracks~\cite{Tang_NeurIPS23_EgoTracks}. The high-quality referring expressions were constructed using a hybrid approach: automatic generation by a Large Language Model (LLM) followed by careful verification and refinement by humans. As illustrated in the comparative analysis in Figure~\ref{fig:refego_longegorefer_comparison}, our benchmark exhibits distinct characteristics that directly address the aforementioned limitations:
1. \textbf{Long-Form Videos}: By leveraging the large-scale egocentric video dataset Ego4D, we target long-form videos averaging 45 minutes in length to evaluate the capability for long-form contextual understanding (Figure~\ref{fig:refego_longegorefer_comparison}a).
2. \textbf{Sparse Appearance}: We intentionally keep the appearance frequency of the referent low (increasing sparsity) to create a more realistic and difficult REC task (Figure~\ref{fig:refego_longegorefer_comparison}b).
3. \textbf{Linguistic Complexity}: As such extensive footage inevitably contains numerous visually ambiguous scenes, we provide highly descriptive referring expressions to uniquely disambiguate the target from similar-looking temporal contexts. This results in significantly increased expression lengths compared to prior benchmarks (Figure~\ref{fig:refego_longegorefer_comparison}c).
4. \textbf{Interaction-Centric}: The majority of our referring expressions are grounded in complex human-object interactions, demanding a high level of scene understanding (Figure~\ref{fig:refego_longegorefer_comparison}, top).

To quantitatively demonstrate the difficulty of our proposed benchmark and to provide a starting point for the research community, we evaluate existing video REC models alongside a training-free baseline we construct. Instead of proposing a definitive end-to-end solution, our training-free baseline modularizes the task into two stages: temporal grounding, leveraging state-of-the-art Vision–Language Models (VLMs) to determine \textit{when} an event occurs, and spatial grounding, utilizing Grounded SAM2~\cite{ren2024grounded} to localize \textit{where} the referred object appears. Our comprehensive experiments reveal a significant performance gap between open-source and closed-source models. Notably, while GPT-5 significantly outperforms all other models, the overall performance of both existing video REC models and our advanced VLM-based baselines remains modest, particularly under stricter evaluation metrics (e.g., mtIoU and mvIoU). This strictly indicates that the task of accurately localizing rare events and objects in long-form videos continues to be a significant and unresolved challenge even for current state-of-the-art models.
The contributions of this research can be summarized as threefold:
\begin{itemize}
\item We introduce \textbf{LongEgoRefer}, a novel and challenging egocentric video REC benchmark. It specifically addresses the limitations of prior work by featuring hour-scale temporal contexts, extreme target sparsity, and complex interaction-centric referring expressions, thereby accurately reflecting realistic first-person visual experiences.
\item We establish standardized baseline strategies to effectively evaluate models on extreme long-form videos. Specifically, we formulate two distinct approaches: (1) adapting existing short-clip video REC models via a sliding window inference strategy, and (2) constructing a training-free, two-stage pipeline that explicitly decouples the task into high-level temporal grounding (via VLMs) and low-level spatial tracking (via Grounded SAM2).
\item We provide a comprehensive empirical analysis exposing the fundamental bottlenecks of current paradigms. We reveal that existing video REC models applied in a sliding-window manner lack global temporal awareness. Meanwhile, the two-stage pipeline suffers from critical error propagation, where inaccuracies in identifying when an event occurs strictly constrain localizing where it happens.
\end{itemize}

\section{Related Work}
\label{sec:related_work}

\subsection{Video-based REC Datasets}
Video-based REC extends image-based grounding~\cite{kazemzadeh2014referit,yu2016modeling,Plummer_2015_ICCV,mao2016generation,Rohrbach_2016_ECCV} to the spatio-temporal domain, aiming to localize target objects in videos based on natural language queries.
Early video REC studies primarily adopted short, trimmed clips (3--10 seconds) and focused on detecting objects or events within brief frame sequences~\cite{li2017cvpr,Xu_2018_ECCV,khoreva2018vos,zhang2020does,Seo2020URVOS}.
Concurrent work explored dynamic human activities and object interactions in third-person videos~\cite{shang2017video,gavrilyuk2018actor,tang2021human,shang2021video,zhang2020does}.
Recent research has moved toward egocentric viewpoints to capture everyday activities.
MeViS~\cite{Ding_2023_ICCV} scales referring video object segmentation to multi-event, motion-centric scenarios in third-person videos.
RefEgo~\cite{kurita2023refego} introduces first-person REC over short clips lasting about 12 seconds on average, with temporal dependencies.
EgoMask~\cite{Liang_2025_ICCV_EgoMask} benchmarks fine-grained spatio-temporal grounding in egocentric videos but covers segments of about 90 seconds with frequently appearing targets.
However, these benchmarks are limited in temporal scale and do not address the challenge of localizing rare object occurrences across extended temporal contexts.
LongEgoRefer advances this direction by introducing hour-scale first-person recordings with sparse target appearances, requiring models to maintain long-term memory while performing precise spatial grounding.

\subsection{Egocentric Video Benchmarks}

Egocentric video benchmarks are essential for understanding human activities from a first-person perspective, directly relevant to embodied AI and human-robot interaction.
Large-scale wearable-camera datasets~\cite{Damen_2018_ECCV,sener2022assembly101,liu2022hoi4d,grauman2022ego4d} have captured daily activities with multi-modal data including RGB, depth, audio, and gaze.
Recent benchmarks further incorporate multi-view synchronization and 3D scene geometry~\cite{Perrett2025_HD-EPIC,Wang2024_EgoVid-5M,Pan_ICCV_AriaDigitalTwin,Wang_ICCV23_HoloAssist,grauman2024ego}.
Building on these datasets, particularly Ego4D, language-conditioned egocentric understanding has emerged across tasks such as video summarization~\cite{Chandrasegaran_NeurIPS24_HourVideo, XLeBench_2025}, video question answering~\cite{mangalam2023egoschema, Baermann_2022_QAEgo4D, Di_2024_CVPR}, and video visual grounding~\cite{ramakrishnan2023naq, kurita2023refego, Liang_2025_ICCV_EgoMask}.
While these works have made significant progress, existing grounding benchmarks focus on short, densely annotated clips where target objects appear frequently and predictably.
In contrast, real-world egocentric videos contain sparse, irregular object occurrences distributed across hours of continuous recording.
LongEgoRefer addresses this gap by requiring models to perform spatio-temporal grounding in hour-long videos where targets may appear only briefly and unpredictably, better reflecting realistic egocentric scenarios.

\subsection{Long-form Video Understanding}
Understanding long-form videos is critical for this work, as it requires models to integrate information across extended time spans rather than within isolated clips.
Early video understanding focused on short, trimmed clips (10--100 seconds) with localized reasoning~\cite{xu2017video,Jang_CVPR17_TGIF-QA,yu_AAAI19_activitynetqa}.
Subsequent datasets increased narrative complexity but remained limited to several minutes~\cite{Xiao_CVPR21_NExT-QA,Li_CVPR24_MVBench,rawal2024cinepile}.
Recent benchmarks extend to long-form comprehension spanning tens of minutes to hours~\cite{mangalam2023egoschema,Song_2024_CVPR,wu2024longvideobench,Chandrasegaran_NeurIPS24_HourVideo,Zhou_CVPR25_MLVU,Fu_CVPR25_Video-MME,Tan_AAAI25_ALLVB}, emphasizing multi-event reasoning and cross-scene consistency.
However, these works primarily address event-level temporal reasoning without spatial grounding.
In contrast, LongEgoRefer requires models to localize rare objects across extended temporal contexts while maintaining precise frame-level spatial grounding.
This setting challenges models to combine long-term memory for tracking rare occurrences with accurate language–visual alignment for fine-grained localization in long-form videos.

\section{Benchmark}
\label{sec:benchmark}

\subsection{Task}
We address the task of \emph{Spatio–Temporal Grounding of Object Occurrences} in long-form egocentric videos.
Given an untrimmed video $V$ of length $T_V$ and a natural language query $Q$, the model must localize a single, continuous appearance of the referred object within the untrimmed video. 
The prediction is represented as a tuple $(t_{\text{start}}, t_{\text{end}}, \mathcal{B})$, where $[t_{\text{start}}, t_{\text{end}}]$ denotes the temporal boundaries satisfying $0 \le t_{\text{start}} \le t_{\text{end}} \le T_V$, and $\mathcal{B}$ is the sequence of bounding boxes corresponding to that duration. 
The key challenge is to find a short, language-relevant moment within hours of egocentric footage and precisely localize the target object throughout that interval, jointly testing temporal localization and spatial grounding under realistic, first-person conditions.

\subsection{Benchmark Creation}
\label{subsec:benchmark creation}
\begin{figure*}[t]
\centering
\includegraphics[width=\linewidth]{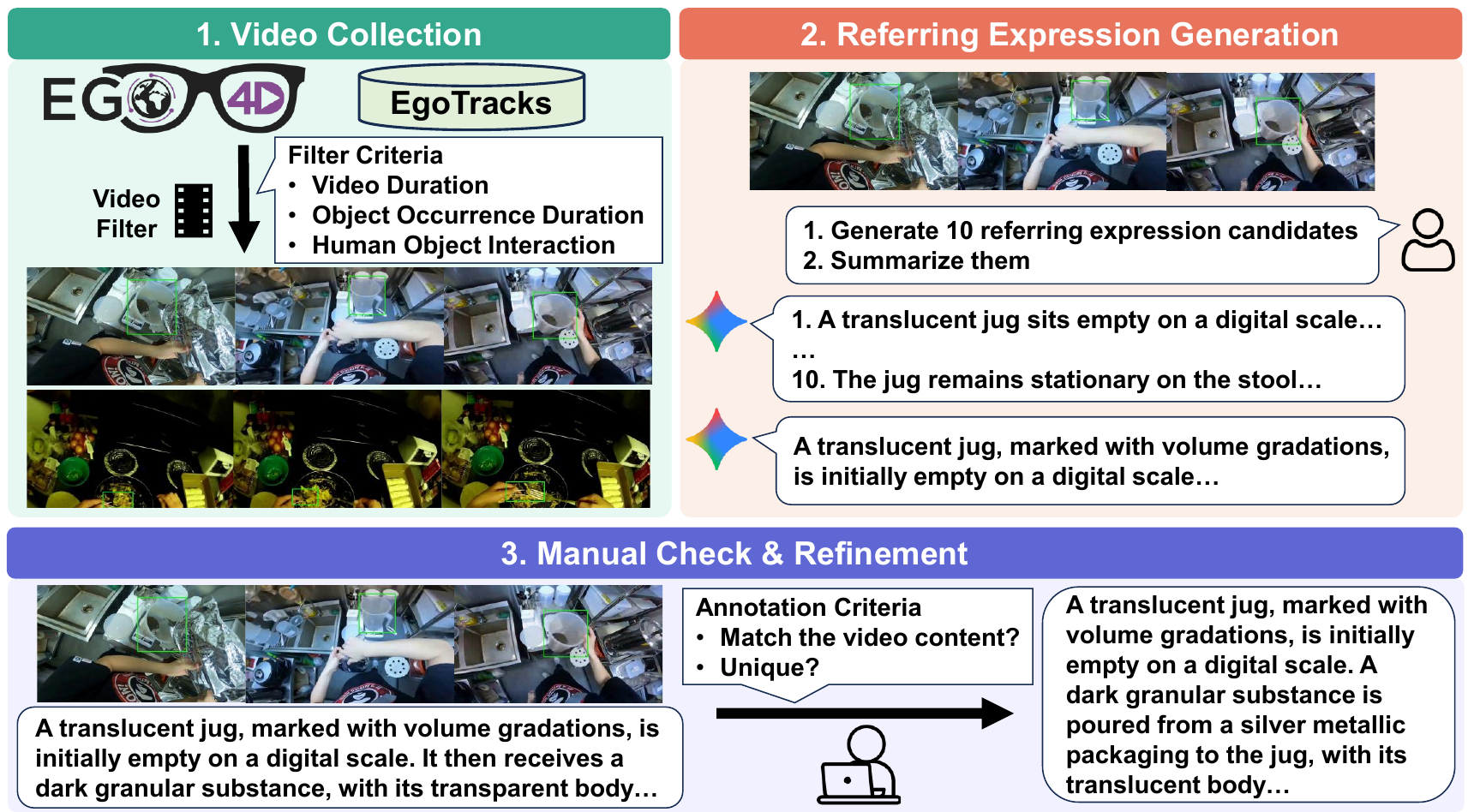}
\caption{An overview of our benchmark construction pipeline.}
\label{fig:benchmark_construction_pipeline}
\end{figure*}

\paragraph{Dataset and annotation preparation}
We build our benchmark upon the annotations from the EgoTracks dataset~\cite{Tang_NeurIPS23_EgoTracks}, which provides long-form object tracking annotations on the large-scale egocentric video corpus Ego4D~\cite{grauman2022ego4d}.
EgoTracks contains per-frame bounding boxes for each object occurrence, along with instance-level attribute information.
While EgoTracks primarily consists of six-minute clips extracted from the original Ego4D videos, our benchmark instead utilizes the \emph{full-length} Ego4D videos to evaluate spatio–temporal grounding under truly long-form conditions.
Although six-minute clips are longer than those in prior grounding datasets, untrimmed hour-scale videos are crucial to evaluate robustness to rare events and long-term temporal dependencies in realistic scenarios.
Importantly, EgoTracks retains timestamp metadata linking each clip to its position in the original video, allowing us to accurately remap existing annotations back to the full-length Ego4D sequences.

\paragraph{Annotation Pipeline}
To ensure the quality, diversity, and uniqueness of our referring expressions, we designed the semi-automated annotation pipeline illustrated in Figure~\ref{fig:benchmark_construction_pipeline}, which consists of a data selection stage and a two-stage language generation process that guarantees each expression does not match other parts of the long video.

\noindent \textbf{(i) Data filtering.} 
To capture realistic long-term scenarios, we first perform strict filtering on the Ego4D dataset. 
We exclude all videos shorter than 20 minutes to ensure sufficient temporal context for long-form reasoning. 
To avoid trivial segments, we retain only object tracks lasting at least 10 seconds where the camera wearer actively manipulates the object (annotated as $\mathtt{is\_active}=\mathtt{True}$ in EgoTracks).
This criterion prioritizes dynamic, interaction-rich episodes that demand temporal grounding.

\noindent \textbf{(ii) Referring expression generation.} 
For each filtered clip, we generate referring expressions using Gemini 2.5 Flash through a two-stage process.
In the first stage, the model is prompted to produce 10 detailed captions describing the target object’s interactions, state changes, and spatial relations.
In the second stage, these captions are aggregated and refined into a single, natural referring expression that captures the temporal context.
This step removes redundancy and produces a refined, cohesive referring expression that encapsulates the temporal context.


\paragraph{Human annotation}

All generated expressions undergo rigorous human verification to ensure their quality and correctness.
Because automatically generated referring expressions can be semantically inconsistent or temporally ambiguous (that is, applicable to multiple moments in the video), we adopt a four-stage human curation pipeline conducted by expert annotators. As a result of this meticulous process, our annotators substantially revised 95.39\% of the Gemini-initialized descriptions to precisely match the video content and ensure referential exclusivity. This extensive human intervention effectively mitigated potential model-induced biases, transforming the initial LLM-generated seeds into high-quality, human-curated grounding queries that provide a fair evaluation platform for all VLM families.
Starting from 2,000 candidate triplets consisting of an Ego4D video, an object occurrence with bounding boxes, and a generated referring expression, we perform the following steps:

\noindent \textbf{(i) Validating and filtering.}
We first check the video segment and temporal boundaries of each occurrence. Samples are removed if (a) the corresponding segment is blank or corrupted, or (b) the object remains visible before or after the annotated timestamps, which makes precise temporal grounding unreliable.

\noindent \textbf{(ii) Correcting expressions.}  
For each valid sample, an annotator verifies whether the referring expression accurately describes the target object and its associated actions. If inaccuracies are found, the expression is rewritten to be factually and semantically correct.

\noindent \textbf{(iii) Disambiguating references.}  
The annotator then ensures that the refined expression uniquely identifies the target moment. If the same description fits other segments in the video, additional visual or temporal cues (such as object attributes, spatial relations, or nearby actions) are added until the expression becomes exclusive to the intended instance.

\noindent \textbf{(iv) Quality assurance.}  
A second annotator independently reviews the results from the previous stages, returning 52.20\% of them for re-annotation until they fully pass the strict quality check. Ten percent of all annotations are selected for an additional verification round to ensure consistency and final quality.

Through this rigorous pipeline, 25.1\% of the initial candidates were filtered out, resulting in a final dataset of 1,498 high-quality curated samples.

\begin{table*}[t]
\centering
\resizebox{\linewidth}{!}{%

\begin{tabular}{lcrrrrr}
\toprule
\textbf{Benchmark} & \textbf{HOI~~} & \textbf{Length (s)} & \textbf{Appear (\%)} & \textbf{\#Videos} & \textbf{\#Ref} & \textbf{\#Words} \\
\midrule
\rowcolor{sectiongray}
\textit{Exocentric Benchmark} &&&&&&\\
Lingual OTB99~\cite{li2017cvpr} & \ding{55} & 19.9 & $\sim$100 & 99 & 99 & 5.5 \\
Lingual ImageNet Videos~\cite{li2017cvpr} & \ding{55} & 9.1 & $\sim$100 & 100 & 100 & 5.6 \\
ReferDAVIS-16~\cite{khoreva2018vos} & \ding{55} & 2.9 & $\sim$100 & 50 & 50 & 5.9 \\
ReferDAVIS-17~\cite{khoreva2018vos} & \ding{55} & 2.9 & $\sim$100 & 120 & 120 & 4.7 \\
Refer-Youtube-VOS~\cite{Seo2020URVOS} & \ding{55} & 4.5 & $\sim$100 & 3,978 & 15,009 & 10.0 \\
MeViS~\cite{Ding_2023_ICCV} & \ding{55} & 13.2 & $\sim$100 & 2,006 & 28,570 & 8.5 \\
\midrule
\rowcolor{sectiongray}
\textit{Egocentric Benchmark} &&&&&&\\
EgoMask~\cite{Liang_2025_ICCV_EgoMask} & \ding{55} & 91.8 & 60.3 & 315 & 700 & 15.0 \\
RefEgo$_\mathrm{test}$~\cite{kurita2023refego} & \ding{55} & 11.9 & 75.5 & 537 & 1,317 & 17.5 \\
\rowcolor{ourrow}
{LongEgoRefer} (Ours) 
& \ding{51} & 2,714.9 & 1.3 & 433 & 1,498 & 63.5 \\
\bottomrule
\end{tabular}
}
\caption{Comparison between LongEgoRefer and existing video REC datasets and benchmarks. ``Appear'' is the appearance rate of the referred object in frames. ``\#Ref'' is the number of referring expressions and ``\#Words'' is average word length.}
\label{tab:benchmark_comparison}
\vspace{-3mm}
\end{table*}

\subsection{Statistics}
The proposed LongEgoRefer benchmark provides a more realistic and challenging setting than previous video REC benchmarks.
Table~\ref{tab:benchmark_comparison} compares LongEgoRefer with existing video referring expression and grounding benchmarks. 
Our proposed LongEgoRefer substantially increases both temporal and linguistic complexity compared to existing video grounding benchmarks. 
It contains 433 egocentric videos with average length of 2,714.9 seconds (approximately 45 minutes) and 1,498 referring expressions with an average of 63.5 words per query, which are significantly longer and more descriptive than those in prior datasets. 
While most existing benchmarks capture short segments where the referred object remains visible throughout the clip (with an appearance rate of 60.3\% or higher), 
the target objects in LongEgoRefer appear in only 1.3\% of frames, requiring models to perform long-range temporal search, contextual reasoning, and memory maintenance over extended sequences. 
Figure~\ref{fig:refego_longegorefer_comparison} highlights clear differences between RefEgo and LongEgoRefer.
RefEgo contains short clips with frequent object appearances, whereas LongEgoRefer consists of much longer videos with sparse targets and longer descriptions.
Furthermore, LongEgoRefer explicitly annotates human–object interactions (HOI), creating a challenging setting for spatio-temporal grounding in real-world, continuous egocentric videos.
These differences make LongEgoRefer a comprehensive benchmark for evaluating models' ability to perform robust, fine-grained spatio-temporal grounding over extended egocentric video sequences, bridging research and real-world applications.

\section{Baseline}
\label{sec:baseline}

To comprehensively evaluate the difficulty of the LongEgoRefer benchmark and provide a starting point for future research, we construct and evaluate two types of baselines: existing end-to-end video REC models adapted for long videos, and a training-free two-stage pipeline leveraging modern visual foundation models.

\subsection{Video REC Models}
Existing video REC models are predominantly designed and trained on short, trimmed clips (typically spanning a few seconds). Consequently, they are fundamentally constrained by memory limitations and context window sizes, making them infeasible to process hour-long videos natively.

To evaluate these conventional models on the extensive sequences of LongEgoRefer, we adopt a sliding window inference strategy. Specifically, we partition the untrimmed long video into a sequence of short temporal windows. We then independently feed each window, along with the referring expression, into the end-to-end video REC model. The model generates a sequence of bounding boxes and an associated confidence score for each window. 

The aggregation strategy for the final prediction depends on the output capabilities of the specific model. For models capable of generating confidence scores for their predictions, we select the temporal window and its corresponding spatio-temporal track that yields the highest confidence score across the entire video. For models that do not provide such evaluation metrics, we instead aggregate the results by temporally concatenating the predicted bounding boxes from all windows to form a single continuous sequence. This sliding window approach serves as a straightforward baseline to gauge how well current short-clip REC models generalize to extreme long-form search scenarios.

\subsection{Training-free Two-stage Pipeline}
Spatio-temporal grounding in long-form videos poses significant computational challenges for end-to-end models. To address this without extensive retraining, we construct a hierarchical pipeline that decouples the task into temporal and spatial reasoning. This design leverages a VLM for global temporal reasoning (``when'') and Grounded SAM2~\cite{ren2024grounded} for local spatial localization (``where''). We adopt Grounded SAM2 for the spatial stage because of its SoTA performance on EgoMask~\cite{Liang_2025_ICCV_EgoMask}.

Specifically, we first employ a pre-trained VLM to identify the temporal interval that matches the query. This coarse estimate is then refined into a precise trajectory using Grounded SAM2. We extract the middle frame of the identified clip and use Grounding DINO~\cite{liu2024grounding} to detect the target object and generate an initial bounding box. This box serves as a visual prompt for SAM2~\cite{ravi2025sam2}, which tracks the object across the segment to produce the final spatio-temporal track. This modular approach provides a clear reference point for the LongEgoRefer benchmark.
\section{Experiments}
\label{sec:experiments}

\begin{table*}[t]
\centering
\resizebox{\linewidth}{!}{%
\begin{tabular}{@{}l|@{}rrr|@{}rrrrr@{}}
\toprule
& \multicolumn{3}{c|}{Temporal} & \multicolumn{5}{c}{Spatio-Temporal} \\
Model & \ tIoU@0.1 & tIoU@0.5 & mtIoU & vIoU@0.1 & vIoU@0.5 & mvIoU & mSTIoU & mIoU+n \\
\midrule
\rowcolor{sectiongray}
\textit{Video REC models} & & & & & & & & \\
VideoLISA~\cite{bai2024one} & 0.45 & 0.07 & 0.15 & 0.13 & 0.00 & 0.37 & 0.34 & 4.90 \\
Sa2VA~\cite{sa2va} & 0.60 & 0.00 & 1.33 & 2.99 & 0.00 & 1.72 & 2.03 & 62.75 \\
Grounded SAM2~\cite{ren2024grounded} & 0.47 & 0.00 & 1.28 & 0.33 & 0.00 & 0.67 & 0.51 & 26.99 \\
SAM3~\cite{carion2025sam} & 0.56 & 0.09 & 0.19 & 0.45 & 0.07 & 0.15 & 0.18 & \textbf{98.82} \\
\midrule
\rowcolor{sectiongray}
\textit{Open-source VLMs} & & & & & & & & \\
InternVideo 2.5 Chat 8B~\cite{intenvideo2.5} & 1.34 & 0.27 & 0.48 & 0.46 & 0.07 & 0.18 & 0.21 & 96.34 \\
MiMo-VL 7B RL~\cite{mimovl2025technical} & 1.13 & 0.20 & 0.42 & 0.33 & 0.00 & 0.13 & 0.15 & 92.65 \\
TimeChat~\cite{ren2024timechat} & 2.00 & 0.20 & 0.79 & 0.73 & 0.00 & 0.24 & 0.28 & 93.31 \\
LITA~\cite{huang2024lita} & 2.60 & 0.06 & 1.14 & 1.46 & 0.06 & 0.50 & 0.59 & 85.99 \\
Video-Chat Flash 7B~\cite{videochat} & 3.07 & 0.47 & 1.07 & 1.20 & 0.13 & 0.37 & 0.45 & 82.45 \\
mPLUG-Owl3 7B~\cite{ye2024mplugowl3} & 2.60 & 0.33 & 0.89 & 1.27 & 0.20 & 0.45 & 0.51 & 88.22 \\
LLaVA-OneVision\! 1.5\! 8B~\cite{an2025llava}\! & 1.20 & 0.33 & 0.57 & 0.67 & 0.07 & 0.19 & 0.20 & 80.58 \\
VideoLLaMA3 7B~\cite{videollama3} & 3.34 & 0.20 & 1.20 & 1.54 & 0.07 & 0.49 & 0.63 & 52.85 \\
LongVA 7B DPO~\cite{longva} & 4.21 & 0.53 & 1.30 & 1.54 & 0.20 & 0.51 & 0.61 & 82.51 \\
LongVILA R1 7B~\cite{longvila} & 2.27 & 0.40 & 0.91 & 1.00 & 0.13 & 0.33 & 0.45 & 66.67 \\
InternVL 3.5 8B~\cite{wang2025internvl3} & 4.34 & 0.07 & 1.50 & 2.27 & 0.07 & 0.66 & 0.84 & 15.98 \\
Qwen3-VL 8B~\cite{bai2025qwen3} & 4.74 & 0.93 & 1.73 & 1.94 & 0.47 & 0.83 & 0.93 & 90.28 \\
InternVL 3.5 38B~\cite{wang2025internvl3} & 6.34 & 0.40 & 1.98 & 2.87 & 0.33 & 0.94 & 1.07 & 78.54 \\
Qwen3-VL 32B~\cite{bai2025qwen3} & 4.87 & 1.07 & 1.81 & 2.20 & 0.47 & 0.84 & 0.97 & 89.83 \\
\midrule
\rowcolor{sectiongray}
\textit{Closed-source VLMs} & & & & & & & & \\
Gemini 2.0 Flash~\cite{comanici2025gemini25} & 5.14 & 1.87 & 2.29 & 2.73 & 0.93 & 1.16 & 1.33 & 97.43 \\
Gemini 2.5 Flash~\cite{comanici2025gemini25} & 10.08 & 3.47 & 4.49 & 5.14 & 1.66 & 2.21 & 2.62 & 95.26 \\
Gemini 2.5 Pro~\cite{comanici2025gemini25} & 23.30 & 8.68 & 10.73 & 12.01 & 4.13 & 5.49 & 6.42 & 95.76 \\ 
GPT-4o~\cite{hurst2024gpt} & 19.69 & 6.54 & 8.71 & 10.68 & 3.47 & 4.49 & 5.11 & 94.55\\
GPT-5~\cite{singh2025openai} & \textbf{37.31} & \textbf{12.55} & \textbf{16.19} & \textbf{17.48} & \textbf{6.20} & \textbf{7.89} & \textbf{8.76} & 95.77\\
\bottomrule
\end{tabular}
}
\caption{Experimental results on our LongEgoRefer benchmark. We evaluate a comprehensive set of open- and closed-source VLMs.}
\vspace{-7mm}
\label{tab:main_results}
\end{table*}

\subsection{Experimental Settings}
\paragraph{Metrics}
For the evaluation on LongEgoRefer, we adopt four metrics. tIoU and vIoU are commonly used metrics for video REC. Following RefEgo, we also employ STIoU and IoU+n.

\paragraph{Implementation Details}
We evaluate 4 video REC models, and 14 open-source and 5 closed-source VLMs on LongEgoRefer. The video REC models are VideoLISA~\cite{bai2024one}, Sa2VA~\cite{sa2va}, Grounded SAM2~\cite{ren2024grounded}, and SAM3~\cite{carion2025sam}. The open-source models are InternVideo 2.5 Chat~\cite{intenvideo2.5}, MiMo-VL~\cite{mimovl2025technical}, TimeChat~\cite{ren2024timechat}, LITA~\cite{huang2024lita}, Video-Chat~\cite{videochat}, mPLUG-Owl3~\cite{ye2024mplugowl3}, LLaVA-OneVision~\cite{an2025llava}, Video-LLaMA3~\cite{videollama3}, LongVA~\cite{longva}, LongVILA R1~\cite{longvila}, InternVL 3.5~\cite{wang2025internvl3}, and Qwen3-VL~\cite{bai2025qwen3}. The closed-source models include Gemini-family (2.0 Flash, 2.5 Flash, and 2.5 Pro)~\cite{comanici2025gemini25} and GPT-family (GPT-4o~\cite{hurst2024gpt} and GPT-5~\cite{singh2025openai}). See Appendix for detailed information.

\subsection{Main Results}

Table~\ref{tab:main_results} presents the quantitative evaluation results of video REC models and leading open-source and closed-source VLMs on our proposed LongEgoRefer benchmark. A key finding from our experiments is that even state-of-the-art models struggle with referring expression comprehension in long-form egocentric videos. The generally low scores across temporal and spatio-temporal metrics highlight the inherent difficulty of accurately localizing a specific clip, often lasting only tens of seconds, within a video that spans tens of minutes to an hour, based solely on referring expressions. 

Notably, the poor performance of video REC models highlights that existing approaches are technically incapable of handling LongEgoRefer directly. The sliding window strategy, while a natural baseline, fails because it lacks the global temporal awareness needed to filter out irrelevant segments across an entire hour, leading to high false-positive matches. This underscores the necessity of our two-stage pipeline: by decomposing the task into temporal and spatial grounding, we leverage the VLM’s extensive context window to prune the vast search space. This ``temporal filtering'' allows the video REC model to focus its fine-grained spatial localization on a manageable, contextually relevant clip, which would otherwise be buried in hours of irrelevant data.

Comparisons across model categories reveal a substantial performance gap between open-source and closed-source models. Even relatively large and high-performing open-source models, such as InternVL and Qwen3-VL, remain in the single digits for tIoU@0.1. Furthermore, it is noteworthy that models specifically designed for long video understanding, such as LongVILA, also show limited effectiveness on our benchmark. In contrast, the closed-source models achieve significantly better performance. Notably, GPT-5 sets a new state of the art, with a tIoU@0.1 of 37.31 and an mtIoU of 16.19, outperforming all other models by a wide margin. Importantly, the fact that GPT-5 consistently outperforms the Gemini series, which was used to generate the initial referring expressions, demonstrates that the benchmark does not suffer from a Gemini-specific bias. This result, coupled with the rigorous human curation process where 95.39\% of initial descriptions were substantially revised (see Section~\ref{subsec:benchmark creation}), confirms that LongEgoRefer evaluates objective spatio-temporal reasoning rather than rewarding model-specific linguistic patterns. This advantage extends to spatio-temporal grounding. While most open-source models struggle to produce accurate object tracks (reflected in mSTIoU scores below 1.0), the closed-source models demonstrate stronger, though still developing, capabilities. GPT-5 again leads, achieving an mSTIoU of 8.76 and an mvIoU of 7.89.  Within the Gemini family, Gemini 2.5 Pro performed best, and 2.5 Flash followed, suggesting that both the model scale and architectural sophistication contribute directly to the performance improvements on complex long-form context understanding.

Despite achieving the best results, the absolute performance of GPT-5 remains relatively low, indicating substantial room for improvement. Temporal localization accuracy appears to be the primary limiting factor, constraining subsequent spatio-temporal grounding metrics. Overall, these results empirically demonstrate that LongEgoRefer presents a highly challenging problem, underscoring the need for novel approaches capable of efficiently processing and deeply understanding long-form video content.

\subsection{Analysis}

\begin{figure*}[t]
\centering
\includegraphics[width=\linewidth]{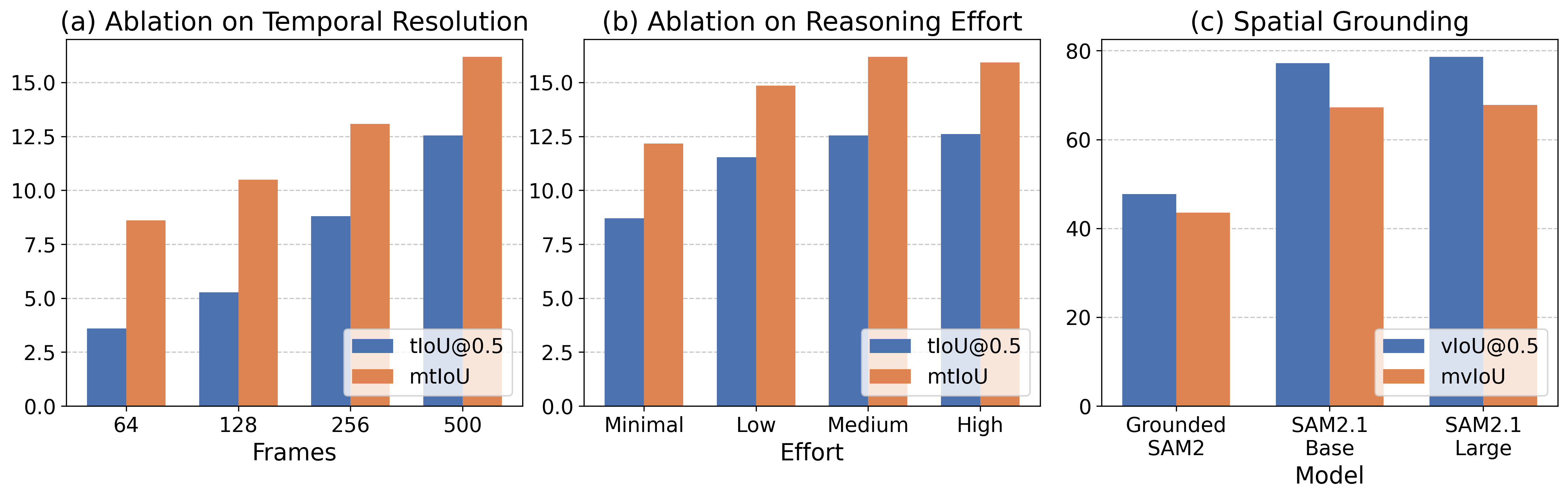}
\caption{Analysis of temporal resolution, reasoning effort, and spatial grounding.}
\label{fig:ablation_graph}
\end{figure*}

\subsubsection{Effect of Temporal Resolution}
Figure~\ref{fig:ablation_graph}a investigates the impact of temporal sampling density on GPT-5 by reducing the maximum frames from the default 500 (constrained by API limits). We observe a significant performance decay as the temporal resolution decreases. Specifically, tIoU@0.5 drops sharply from 12.55 to 3.60 when the input is limited to 64 frames. This suggests that sparse sampling leads to the loss of fine-grained visual cues essential for precise temporal localization in long-form videos. These results confirm that high temporal resolution is critical for accurate grounding and that GPT-5 effectively leverages dense 500-frame contexts to model complex temporal dependencies.

\subsubsection{Effect of Reasoning Effort}
Figure~\ref{fig:ablation_graph}b investigates the impact of reasoning effort levels on GPT-5. Specifically, ``Minimal'' corresponds to zero reasoning effort, whereas ``High'' represents the maximum available reasoning capacity. The results demonstrate that increasing reasoning effort yields substantial performance gains, with tIoU@0.5 improving from 8.7 to 12.61. Notably, we observe that performance largely saturates at the ``Medium'' level, which serves as the API's default setting. This suggests that while more intensive reasoning generally benefits temporal localization, there may be a point of diminishing returns or increased noise.

\subsubsection{Spatial Grounding}

We further investigate the importance of temporal grounding. We evaluate spatial grounding alone with the annotated (oracle) and \allowbreak Grounded-SAM2-based temporal segmentation for the object appearance of each video.
With the oracle temporal segmentation, we evaluate the segmentation model of SAM2.1 Base and Large. We use the ground-truth bounding box in the first frame for the prompts of SAM2.1.
Figure~\ref{fig:ablation_graph}c presents the results in terms of vIoU. Compared to the results in Table~\ref{tab:main_results}, the scores have improved dramatically with the oracle temporal segmentation. For example, while GPT-5 achieves an mvIoU of 7.89 when using its own temporal grounding results, Grounded SAM2 attains 43.54 (+35.65). This demonstrates that the accuracy of temporal grounding has a substantial impact on the overall performance of spatio-temporal grounding.
Furthermore, when comparing Grounded SAM2 with SAM2.1 Base and SAM2.1 Large, mvIoU increases to 67.21 (+23.67) and 67.82 (+24.28), respectively. These results indicate that the precision of the bounding box used as the prompt for SAM2 significantly contributes to the performance of spatial grounding.

\subsubsection{Results on EgoTracks Videos}

\begin{table*}[t]
\centering
\resizebox{\linewidth}{!}{%
\begin{tabular}{@{}l|@{}rrr|@{}rrrrr@{}}
\toprule
& \multicolumn{3}{c|}{Temporal} & \multicolumn{5}{c}{Spatio-Temporal} \\
Model & \ tIoU@0.1\! & tIoU@0.5\! & mtIoU\! & \ vIoU@0.1\! & vIoU@0.5\! & mvIoU\! & mSTIoU\! & mIoU+n\\
\midrule
Gemini 2.0 Flash & 39.38 & 14.35 & 17.70 & 19.49 & 6.40 & 8.44 & 9.78 & \textbf{96.95}  \\
Gemini 2.5 Flash & 45.19& 15.28 & 20.36 & 22.76 & \textbf{8.21} & 10.21 & 11.83 & 95.89  \\
Gemini 2.5 Pro & \textbf{49.93} & \textbf{19.35} & \textbf{22.97} & \textbf{25.43} & 8.07 & \textbf{10.88} & \textbf{12.63} & 96.44  \\ 
\bottomrule
\end{tabular}
}
\caption{Experimental results on EgoTracks videos.}
\vspace{-3mm}
\label{tab:egotracks_results}
\end{table*}

To better understand the challenges of long-form videos in video REC, we evaluate Gemini on LongEgoRefer using the EgoTracks dataset instead of the original Ego4D videos.
EgoTracks is constructed by trimming the Ego4D videos to an average length of six minutes. Although shorter than the original Ego4D videos (45 minutes on average), EgoTracks still consists of considerably longer videos than those in RefEgo or EgoMask.
The results are summarized in Table~\ref{tab:egotracks_results}.
Compared to the results on the original long-form Ego4D videos, the performance improves substantially.
For instance, Gemini 2.5 Pro achieves a gain of +12.24 mtIoU and +5.39 mvIoU.
These results indicate that while long-form video REC is an important and practical problem, it remains highly challenging.
We hypothesize that the main factors behind the performance degradation on long videos are the repeated occurrence of visually similar events and the decreased frequency of target object appearances.

\subsection{Qualitative Results}

\begin{figure*}[t]
\centering
\includegraphics[width=\linewidth]{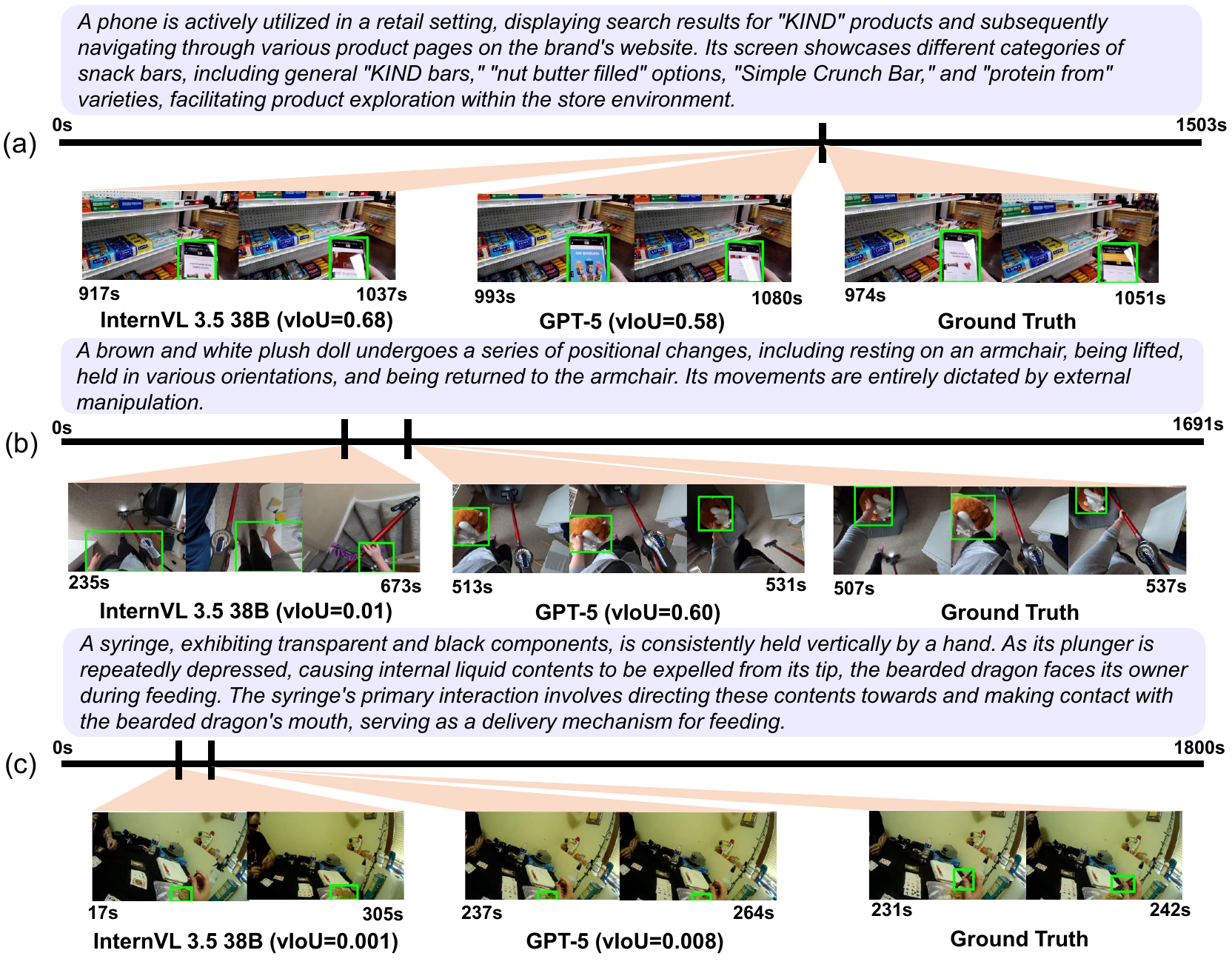}
\caption{Qualitative results.}
\label{fig:qualitative_analysis}
\vspace{-3mm}
\end{figure*}

We present prediction examples of InternVL 3.5 38B, GPT-5, and ground truth in Figure~\ref{fig:qualitative_analysis}. In the example (a), both InternVL and GPT predict the correct bounding boxes. In the example (b), GPT predicts the correct bounding boxes, but InternVL predicts the wrong temporal duration and bounding boxes. In the example (c), both InternVL and GPT predict the wrong bounding boxes, even though the temporal duration of GPT is close to the ground truth.

\section{Conclusion}
\label{sec:conclusion}

In this work, we introduced \textbf{LongEgoRefer}, a new benchmark for egocentric video referring expression comprehension that addresses key limitations of existing benchmarks. It substantially expands the temporal scale (averaging 45 minutes per video) and introduces extreme target sparsity, where referents are visible for only 1.3\% of the video, requiring models to identify brief and infrequent visual occurrences within long, continuous streams.
Furthermore, LongEgoRefer is the first egocentric benchmark explicitly centered on complex human-object interactions (HOI), accompanied by long, descriptive expressions averaging over 63 words, which demand advanced spatio-temporal reasoning and language grounding.
Our experiments show that these characteristics present significant challenges to current vision-language models. Even the strong training-free baselines we constructed, which combine state-of-the-art VLMs with the powerful spatial tracker, achieve limited accuracy. 
LongEgoRefer thus establishes a new standard for real-world complexity and provides a foundation for developing models with robust long-term memory, interaction reasoning, and the ability to ground complex language in sparse and cluttered visual environments, advancing research toward more capable human-centric AI systems.

This study focuses on constructing a rigorous benchmark for spatio-temporal grounding in long-form egocentric videos. 
While LongEgoRefer provides a critical testbed, developing a comprehensive, large-scale training dataset for videos spanning tens of minutes remains an important direction for future work. For practical applications such as human-robot interaction in daily-life scenarios, extending the benchmark to multi-hour or day-long continuous videos will be indispensable.


\section*{Acknowledgements}
This work was supported by JST PRESTO Grant Number JPMJPR22P8, JST K Program, Grant Number JPMJKP25V2, JST CRONOS, Grant Number JPMJCS24K6,  and JST BOOST, Grant Number JPMJBY24C5. This work used the TSUBAME4.0 supercomputer at Institute of Science Tokyo.

%
%
\bibliographystyle{splncs04}
\bibliography{main}

\clearpage
\appendix

\section{Word Cloud}

\begin{figure}[t]
\centering
\includegraphics[width=0.5\linewidth]{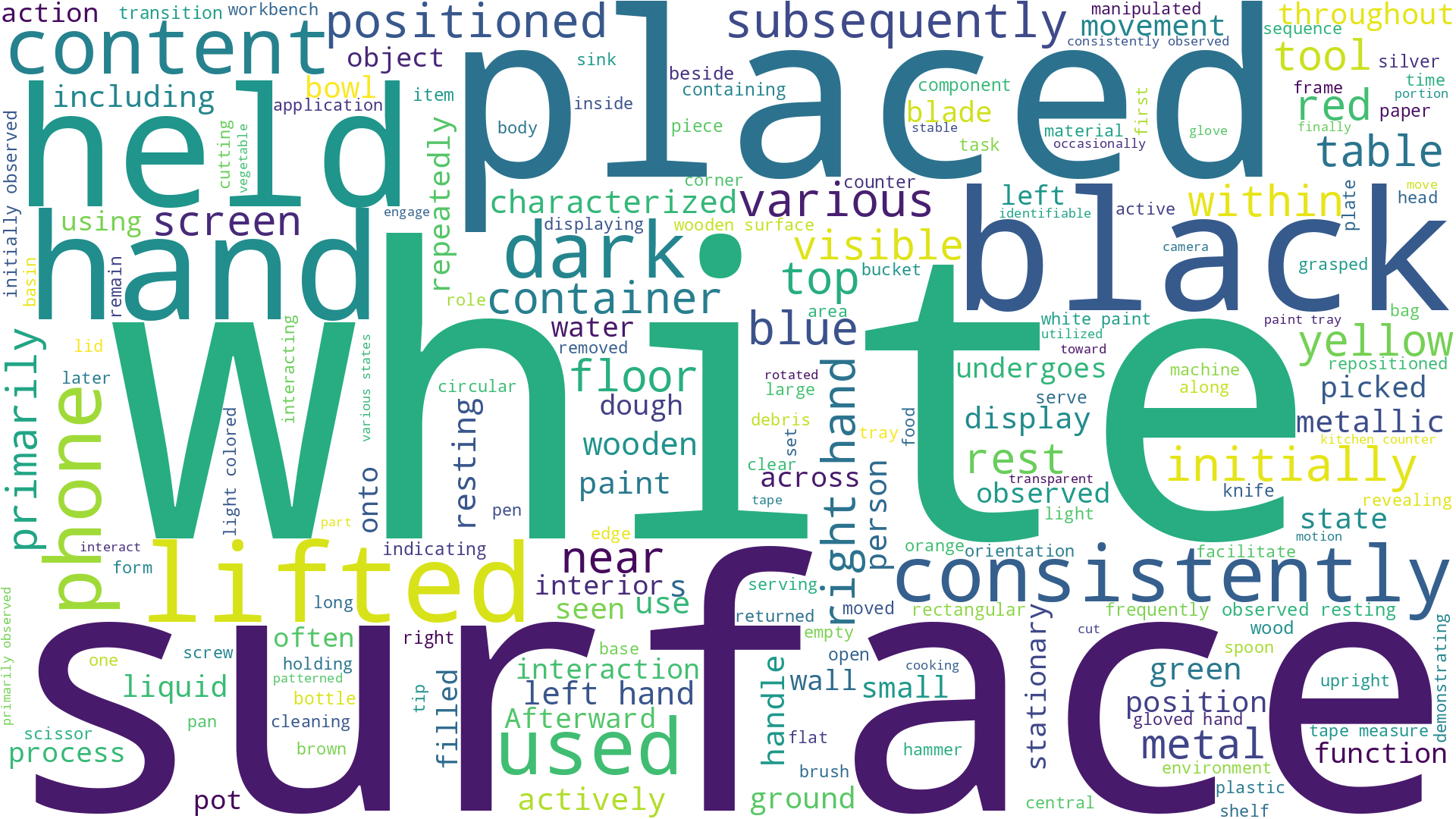}
\caption{Word cloud visualization of referring expressions in LongEgoRefer.}
\label{fig:expression_wordcloud}
\end{figure}

Figure~\ref{fig:expression_wordcloud} visualizes the linguistic diversity of referring expressions in LongEgoRefer. The prominent size of action verbs such as ``placed,'' ``lifted,'' and ``held,'' alongside the noun ``hand,'' illustrates the benchmark's strong emphasis on\\ Human-Object Interaction (HOI). This distribution confirms that objects in egocentric video are primarily identified through dynamic manipulation and their spatial relationship with the user's hands and surfaces. The benchmark also relies heavily on visual attributes (e.g., ``white,'' ``black,'' ``dark'') and spatial context (e.g., ``surface,'' ``positioned''). Furthermore, temporal indicators such as ``subsequently'' and ``initially'' highlight the need for models to track object state changes over time.

\section{Object Distribution}

\begin{figure*}[t]
\centering
\includegraphics[width=\linewidth]{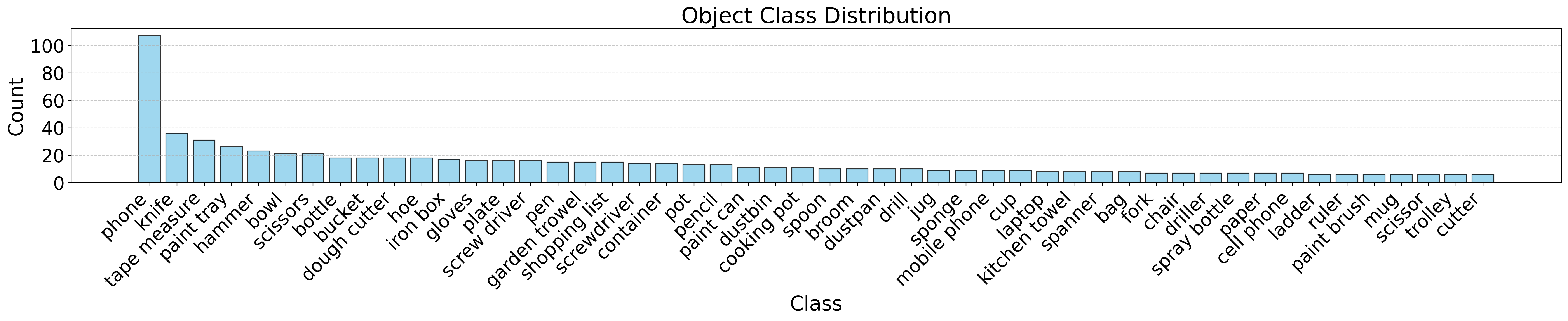}
\caption{Object class distribution from EgoTracks.}
\label{fig:object_distribution}
\end{figure*}

Figure~\ref{fig:object_distribution} presents the object distribution of LongEgoRefer annotated by EgoTracks~\cite{Tang_NeurIPS23_EgoTracks}. As LongEgoRefer is interaction-centric, the most frequent objects are everyday items such as phones and knives, reflecting its practical and application-oriented nature.

\section{Interaction Distribution}

\begin{figure*}[t]
\centering
\includegraphics[width=\linewidth]{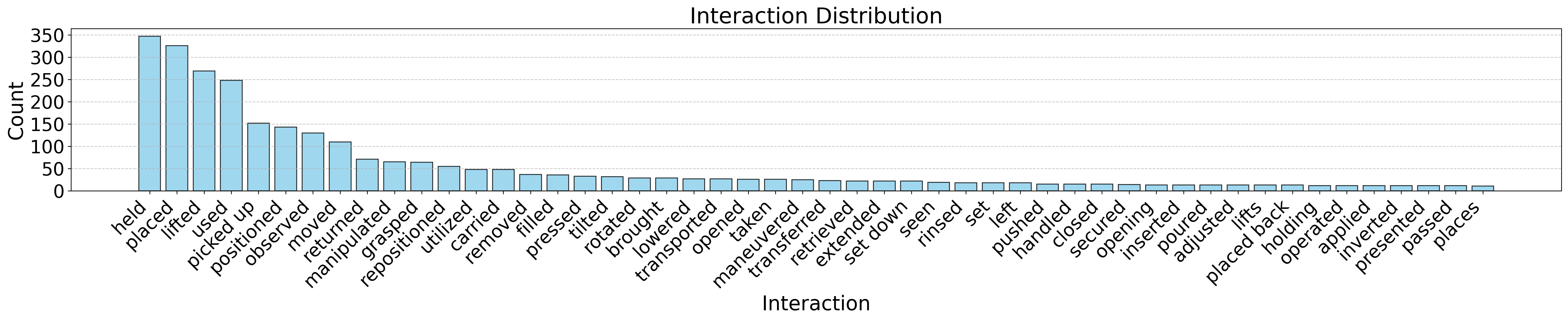}
\caption{Distribution of Human-Object Interactions.}
\label{fig:interaction_distribution}
\end{figure*}

Figure~\ref{fig:interaction_distribution} presents the distribution of Human-Object Interactions in LongEgoRefer. We employ GPT-5 to extract interaction verbs from the referring expressions in our benchmark. The results highlight the linguistic and semantic richness of the dataset, covering a diverse range of egocentric activities.

\section{EgoTracks Split and Annotation Cost}

We use the train and validation splits of EgoTracks for LongEgoRefer benchmark construction. The total annotation cost amounts to \$4,700. All 1,498 curated queries are designed exclusively as an evaluation benchmark (test set). As modern foundation models and VLMs are increasingly evaluated on their zero-shot generalization capabilities, we do not provide a specific training split. Instead, LongEgoRefer serves as a pure testbed for zero-shot long-form spatio-temporal reasoning. 

\section{Training-Free Two-Stage Pipeline}
\begin{figure}[t]
\centering
\includegraphics[width=0.5\linewidth]{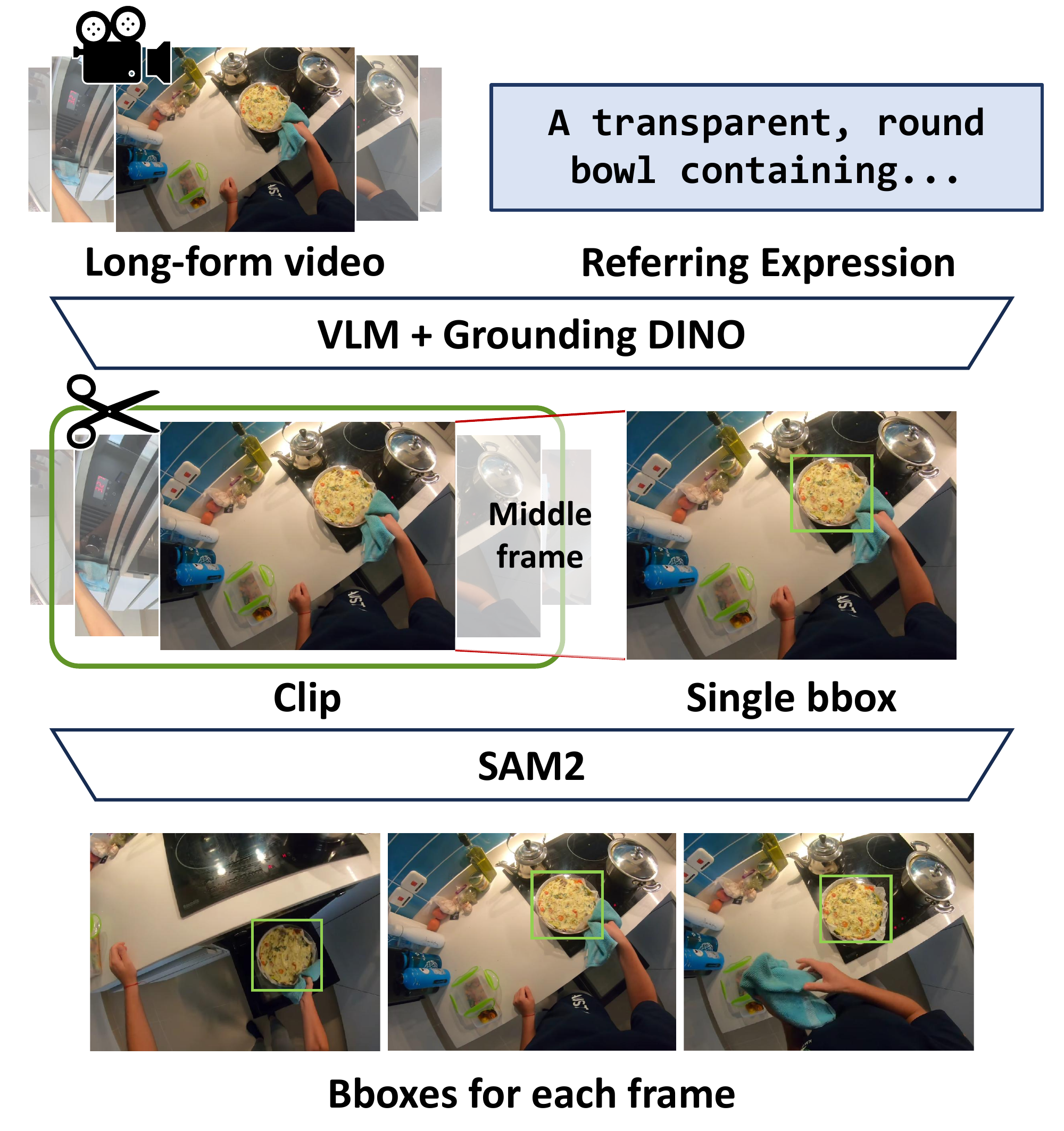}
\caption{Our training-free baseline.}
\label{fig:baseline}
\end{figure}

Our baseline adopts a two-stage pipeline that sequentially performs temporal and spatial grounding, as illustrated in Figure~\ref{fig:baseline}.
\paragraph{Stage 1: Temporal Grounding}
The primary objective of this stage is to identify the temporal interval $T = [t_{\text{start}}, t_{\text{end}}]$ within a potentially long and untrimmed video $V$ that semantically corresponds to a given textual query $Q$. 
To achieve this, we leverage a pre-trained VLM, capitalizing on its advanced capabilities for processing and reasoning over extended video sequences.

We formulate the temporal grounding task as a structured generation problem. 
The VLM must output the start time $t_{\text{start}}$ and end time $t_{\text{end}}$ that define the temporal boundaries of the described event.

This process can be formally expressed as:
\begin{equation}
    t_{\text{start}}, t_{\text{end}} = f_{\text{VLM}}(V, Q),
\end{equation}
where $f_{\text{VLM}}$ represents the prompted VLM. 
A key advantage of this approach is the VLM's inherent ability to process long contexts. This allows the baseline to ground queries that may refer to events, objects, or their relationships separated by large temporal distances within the video. 

\paragraph{Stage 2: Spatial Grounding}
The second stage refines the coarse temporal estimate from the VLM into a precise spatio-temporal track of the object. 
For this purpose, we employ Grounded-SAM2, a two-step pipeline consisting of \textbf{Grounding DINO}~\cite{liu2024grounding} and \textbf{SAM2}~\cite{ravi2025sam2}, which are renowned for their exceptional zero-shot, prompt-guided object detection and segmentation capabilities, respectively.

First, we extract the clip $C$ corresponding to the timestamp $[t_{\text{start}}, t_{\text{end}}]$ identified by the VLM. 
To initialize the track, we provide the original referring expression $Q$ and the \textbf{middle frame} of the clip, $I_{\text{mid}}$, which is likely to contain a representative view of the target object, to Grounding DINO. 
This generates a high-quality initial bounding box for the target object, which we will use as a prompt:
\begin{equation}
    B_{\text{prompt}} = f_{\text{GD}}(I_{\text{mid}}, Q),
\end{equation}
where $f_{\text{GD}}$ denotes the Grounding DINO model and $I_{\text{mid}}$ is the frame at timestamp $(t_{\text{start}} + t_{\text{end}}) / 2$.

This initial bounding box, $B_{\text{prompt}}$, is then used as a spatial prompt to guide SAM2 to track the object throughout the entire clip $C$. 
The operation is formulated as:
\begin{equation}
    \mathcal{B} = f_{\text{SAM2}}(C, B_{\text{prompt}}),
\end{equation}
where $f_{\text{SAM2}}$ denotes the SAM2 model operating in video tracking mode, and $\mathcal{B}$ is the final bounding box sequence of the target object.

This two-stage design creates a modular workflow where the VLM handles the high-level \textit{when} and Grounded-SAM2 handles the low-level \textit{where}. While this decomposition is conceptually intuitive, it inherently faces the challenge of error propagation, where Stage 1 inaccuracies inevitably constrain Stage 2 performance. By evaluating this pipeline, we aim to expose the fundamental difficulties of the LongEgoRefer benchmark and provide a clear baseline for future work to improve upon, particularly in bridging the gap between temporal and spatial reasoning.

\section{Metric}

For the evaluation on LongEgoRefer, we adopt four base metrics. tIoU and vIoU are commonly used metrics for video REC. Following RefEgo~\cite{kurita2023refego}, we also employ STIoU and IoU+n, as LongEgoRefer contains a large number of background frames without target objects.
\begin{itemize}
    \item \textbf{mtIoU}: This metric evaluates the temporal localization accuracy. It is computed as the Intersection over Union (IoU) between the predicted temporal segment and the ground-truth temporal segment.
    \item \textbf{tIoU@R}: This metric, also known as localization recall or success rate, measures the percentage of queries for which the tIoU exceeds a certain threshold $R$. It provides a binary measure of successful temporal localization. Following standard practice, we report the recall for multiple thresholds, $R \in \{0.1, 0.5\}$.
    \item \textbf{mvIoU}: This metric assesses the spatio-temporal localization accuracy by measuring the overlap between a sequence of predicted and ground-truth bounding boxes. It is calculated by averaging the spatial IoU of the bounding boxes over the frames where both the predicted and ground-truth objects are present.
    \item \textbf{vIoU@R}: Similar to tIoU@R, this metric evaluates the percentage of samples where the vIoU score is above a threshold $R$. This serves as a strict measure of successful spatio-temporal localization. We report the recall for multiple thresholds, $R \in \{0.1, 0.5\}$.
    \item \textbf{mSTIoU}: This metric is adopted in RefEgo~\cite{kurita2023refego}. While vIoU is based on frame-by-frame IoU, STIoU is a multi-frame-based IoU.
    \item \textbf{mIoU+n}: This metric adopted in RefEgo is similar to mvIoU, but IoU is set to 1 for frames that are true negatives (i.e., when the frame is negative in both the prediction and ground truth).
\end{itemize}

\section{Prompt}

\begin{figure*}[t]
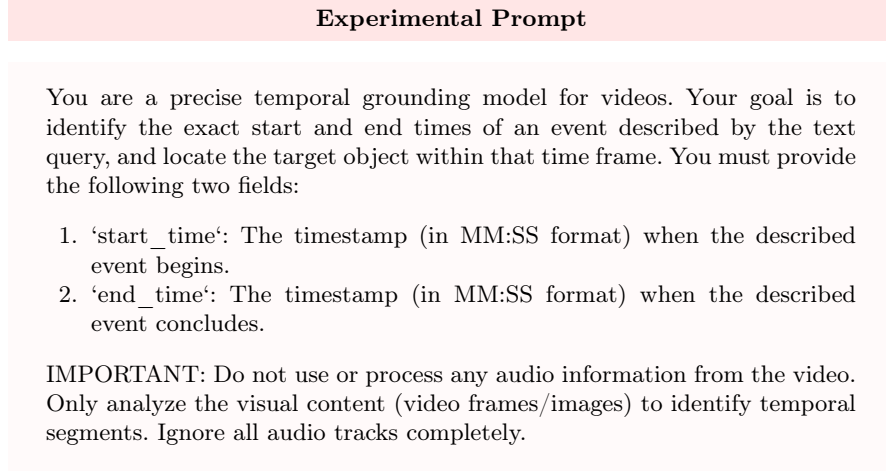

\centering
\begin{minipage}{0.96\textwidth}

\begin{mdframed}[
  backgroundcolor=red!10,
  linecolor=white,
  linewidth=0pt,
  innerleftmargin=10pt,
  innerrightmargin=10pt,
  innertopmargin=4pt,
  innerbottommargin=4pt,
  skipabove=0pt,
  skipbelow=0pt
]
\begin{center}
\small
\textbf{Experimental Prompt}
\end{center}
\end{mdframed}%

\begin{mdframed}[
  backgroundcolor=red!2,
  linecolor=white,
  linewidth=0pt,
  innerleftmargin=14pt,
  innerrightmargin=14pt,
  innertopmargin=10pt,
  innerbottommargin=10pt,
  skipabove=0pt,
  skipbelow=0pt
]
\footnotesize

You are a precise temporal grounding model for videos.
Your goal is to identify the exact start and end times of an event described by the text query, and locate the target object within that time frame.
You must provide the following two fields:
\begin{enumerate}
    \item `start\_time`: The timestamp (in MM:SS format) when the described event begins.
    \item `end\_time`: The timestamp (in MM:SS format) when the described event concludes.
\end{enumerate}

IMPORTANT: Do not use or process any audio information from the video. Only analyze the visual content (video frames/images) to identify temporal segments. Ignore all audio tracks completely.

\end{mdframed}
\end{minipage}

\caption{Prompt used in experiments.}
\label{fig:referring_experiment_prompt}
\end{figure*}

We present prompts for referring expression generation in Figure~\ref{fig:referring_expression_generation_prompt} and the experimental prompts in Figure~\ref{fig:referring_experiment_prompt}.

\begin{figure*}[t]
\centering
\begin{minipage}{0.96\textwidth}

\begin{mdframed}[
  backgroundcolor=red!10,
  linecolor=white,
  linewidth=0pt,
  innerleftmargin=10pt,
  innerrightmargin=10pt,
  innertopmargin=4pt,
  innerbottommargin=4pt,
  skipabove=0pt,
  skipbelow=0pt
]
\begin{center}
\small
\textbf{Referring Expression Generation Prompt}
\end{center}
\end{mdframed}%

\begin{mdframed}[
  backgroundcolor=red!2,
  linecolor=white,
  linewidth=0pt,
  innerleftmargin=14pt,
  innerrightmargin=14pt,
  innertopmargin=10pt,
  innerbottommargin=10pt,
  skipabove=0pt,
  skipbelow=0pt
]
\footnotesize

You are tasked with generating detailed descriptions of the target object in the given video clip, focusing on its interactions with the camera wearer.

A green bounding box highlights the target object in each frame to assist in identification. Each description must be object-centric, starting with the object's name and emphasizing its attributes, movements, and interactions. Avoid human-centric descriptions.

Example:
\begin{itemize}
    \item \checkmark Object-centric: A white coffee cup is held in a person's right hand as they sip from it while standing near a kitchen counter. The kitchen features wooden cabinets and a small window in the background.
    \item $\times$ Human-centric (do not generate): A person is holding a white coffee cup in their right hand, sipping from it while standing near a kitchen counter. The kitchen features wooden cabinets and a small window in the background.
\end{itemize}

The target object name is \{object class\}. You must generate 10 unique descriptions, ensuring each is distinct.

Final Task: Object-Centric Summary
After generating the descriptions, provide a concise, object-centric summary that highlights the key subjects, attributes, and interactions of the target object. The summary should:
\begin{itemize}
    \item Focus on objects, avoiding human-centric phrasing.
    \item Capture the roles, attributes, and significance of the target object.
    \item Be clear, coherent, and representative of the descriptions.
\end{itemize}

Output Format:

\verb|```|

DESCRIPTION: ``description 1'',

DESCRIPTION: ``description 2'',

...

DESCRIPTION: ``description 10'',

SUMMARY: ``summary of descriptions''

\verb|```|

\end{mdframed}
\end{minipage}

\caption{Referring Expression Generation Prompt.}
\label{fig:referring_expression_generation_prompt}
\end{figure*}

\section{Implementation Details}
For video REC models, we aggregate predictions over 1-minute video clips. Among these models, only SAM3 is capable of predicting confidence scores for its outputs, while the other models do not provide such scoring metrics.
For Gemini, videos are sampled at 1 FPS. Each frame is converted into 258 tokens. When video durations exceed 1 hour, we enable the low-resolution mode, where each frame is represented by 66 tokens. For GPT, videos are temporally sampled at equal intervals to comply with the API limits, and up to 500 frames are extracted. Each frame is resized while preserving the aspect ratio so that the longer side is 512 pixels. For open-source VLMs, frames with a shorter edge of approximately 400 pixels are input up to 128 frames or the maximum window size of each model. As VLMs sometimes predict very long temporal duration, such as over 1,000 seconds, we set the maximum temporal duration in the spatio-temporal evaluation as 300 seconds to reduce the computational cost. We use two NVIDIA H100 GPUs.

\section{More Analysis}
\paragraph{Effect of Video Duration}


Table~\ref{tab:effect_of_video_duration} presents experimental results of mvIoU on LongEgoRefer with respect to the video duration. We split the duration into three ranges: 20-40 minutes, 40-60 minutes, and 60-120 minutes. The results demonstrate that for most models, including GPT-5, the performance is highest in the relatively shorter 20-40 minutes range and drops as the video length increases. This performance degradation is likely due to the strict limit on the maximum number of input frames (e.g., 500 frames for GPT-5 and 128 frames for open-weight VLMs), which significantly reduces the temporal resolution for longer videos. Interestingly, Gemini 2.5 Pro exhibits a distinct trend, achieving its best performance in the 60-120 minutes range. We attribute this robustness to Gemini's massive context window, which allows it to maintain a dense sampling rate of 1 FPS regardless of the video duration. By preserving fine-grained temporal information even in extreme long-form videos, Gemini effectively avoids the temporal information loss that severely hinders other models.

\begin{table*}[t]
\centering
\begin{tabular}{@{}l|@{}rrr@{}}
\toprule
& \multicolumn{3}{c}{mvIoU} \\
Model & \ (20, 40] & (40, 60] & (60, 120] \\
\midrule
InternVL 3.5 38B~\cite{wang2025internvl3}& 1.18 & 0.53 & 0.90 \\
Qwen3-VL 32B~\cite{bai2025qwen3} & 1.43 & 0.21 & 0.57 \\
Gemini 2.5 Pro~\cite{comanici2025gemini25} & 5.50 & 3.85 & \textbf{7.33}  \\ 
GPT-5~\cite{singh2025openai} & \textbf{9.46} & \textbf{5.86} & 6.95  \\ 
\bottomrule
\end{tabular}
\caption{Experimental results on our LongEgoRefer benchmark with respect to the video duration.}
\label{tab:effect_of_video_duration}
\end{table*}

\paragraph{Effect of Appearance Rate}


Table~\ref{tab:effect_of_appearance_rate} presents experimental results of mvIoU on LongEgoRefer with respect to the target object's appearance rate. We split the rate into three ranges: 0-1\%, 1-2\%, and 2-25\%. The results reveal a stark correlation between the appearance rate and the localization performance. All evaluated models struggle significantly when the object appears sparsely within the video. For instance, even the state-of-the-art GPT-5 and Gemini 2.5 Pro experience a drastic performance drop of over 50\% when the appearance rate falls below 1\% compared to the >2\% range, while open-weight models fail almost entirely. This confirms that precisely localizing rare objects remains a fundamental bottleneck for current VLMs, highlighting the rigorous nature of our benchmark.

\begin{table*}[t]
\centering
\begin{tabular}{@{}l|@{}rrr@{}}
\toprule
& \multicolumn{3}{c}{mvIoU} \\
Model & \ (0, 1] & (1, 2] & (2, 25] \\
\midrule
InternVL 3.5 38B~\cite{wang2025internvl3}& 0.45 & 1.12 & 2.67 \\
Qwen3-VL 32B~\cite{bai2025qwen3} & 0.40 & 1.17 & 2.49 \\
Gemini 2.5 Pro~\cite{comanici2025gemini25} & 4.05 & 7.07 & 8.78  \\ 
GPT-5~\cite{singh2025openai} & \textbf{5.16} & \textbf{11.22} & \textbf{13.54}  \\ 
\bottomrule
\end{tabular}
\caption{Experimental results on our LongEgoRefer benchmark with respect to the appearance rate.}
\label{tab:effect_of_appearance_rate}
\end{table*}

\paragraph{Effect of Referring Expression Length}


Table~\ref{tab:effect_of_referring_expression} presents experimental results of mvIoU on LongEgoRefer with respect to the referring expression length. We split the length into three ranges: 0-50 words, 50-70 words, and 70-110 words. The results demonstrate that linguistic complexity has a significant impact on the difficulty of the task. Specifically, as the referring expression becomes longer, the performance of the VLMs, including GPT-5 and Gemini 2.5 Pro, consistently degrades. Longer expressions typically contain more complex constraints, such as detailed visual attributes, fine-grained object states, and sequential interactions, which challenge the models' ability to align extended linguistic queries with long-form visual contexts. Notably, while GPT-5 maintains the best performance across all lengths, its score drops by over 40\% from the shortest to the longest queries, highlighting the necessity for stronger fine-grained multimodal alignment in future foundation models.

\begin{table*}[t]
\centering
\begin{tabular}{@{}l|@{}rrr@{}}
\toprule
& \multicolumn{3}{c}{mvIoU} \\
Model & \ (0, 50] & (50, 70] & (70, 110] \\
\midrule
InternVL 3.5 38B~\cite{wang2025internvl3} & 1.02 & 0.92 & 0.91 \\
Qwen3-VL 32B~\cite{bai2025qwen3} & 0.81 & 1.14 & 0.06 \\
Gemini 2.5 Pro~\cite{comanici2025gemini25} & 7.96 & 5.85 & 3.62  \\ 
GPT-5~\cite{singh2025openai} & \textbf{10.29} & \textbf{8.25} & \textbf{6.05}  \\ 
\bottomrule
\end{tabular}
\caption{Experimental results on our LongEgoRefer benchmark with respect to the referring expression length.}
\label{tab:effect_of_referring_expression}
\end{table*}

\paragraph{Effect of Temporal Position}

To further investigate the impact of event timing on grounding performance, we analyzed the mvIoU with respect to the temporal position of the target event within long-duration videos. We divided each video into three segments: early (0, 1/3], middle (1/3, 2/3], and late (2/3, 1]. As shown in Table~\ref{tab:effect_of_temporal_distance}, compared to the early segment, all evaluated models exhibit a significant performance drop when the target event occurs in the middle or late segments of the video.

\begin{table*}[t]
\centering
\begin{tabular}{@{}l|@{}rrr@{}}
\toprule
& \multicolumn{3}{c}{mvIoU} \\
Model & \ (0, 1/3] & (1/3, 2/3] & (2/3, 1] \\
\midrule
InternVL 3.5 38B~\cite{wang2025internvl3} & 1.50 & 0.65 & 0.66 \\
Qwen3-VL 32B~\cite{bai2025qwen3} & 1.59 & 0.51 & 0.59 \\
Gemini 2.5 Pro~\cite{comanici2025gemini25} & 6.94 & 4.60 & 4.95  \\ 
GPT-5~\cite{singh2025openai} & \textbf{9.61} & \textbf{7.79} & \textbf{6.17}  \\ 
\bottomrule
\end{tabular}
\caption{Effect of temporal position on LongEgoRefer.}
\label{tab:effect_of_temporal_distance}
\end{table*}

\paragraph{Effect of Object Scale}

Table~\ref{tab:effect_of_object_scale} presents experimental results of mvIoU on LongEgoRefer with respect to the object scale. As shown in the table, both models exhibit a consistent trend where performance scales with the size of the target object. For GPT-5, mvIoU increases from 5.03 for small objects to 10.76 for large objects, more than doubling across the scale spectrum. These results suggest that detecting and tracking smaller objects remains a significant challenge for existing VLMs, likely due to the limited visual resolution and information density of smaller regions of interest in egocentric videos.

\begin{table*}[t]
\centering
\begin{tabular}{@{}l|@{}rrr@{}}
\toprule
& \multicolumn{3}{c}{mvIoU} \\
Model &  Small & Medium & Large \\
\midrule
Gemini 2.5 Pro~\cite{comanici2025gemini25} & 3.43 & 5.17 & 7.80  \\ 
GPT-5~\cite{singh2025openai} & \textbf{5.03} & \textbf{7.79} & \textbf{10.76}  \\ 
\bottomrule
\end{tabular}
\caption{Effect of object scale on LongEgoRefer.}
\label{tab:effect_of_object_scale}
\end{table*}

\paragraph{Effect of Velocity}

Table~\ref{tab:effect_of_velocity} presents experimental results of mvIoU on LongEgoRefer with respect to the target's velocity. We observe an inverse relationship between performance and object velocity. For GPT-5, mvIoU drops from 10.94 for slow-moving objects to 4.84 for fast-moving objects. The fast-moving objects often introduce motion blur and rapid spatial displacement between frames, which appears to substantially hinder the models' ability to maintain accurate temporal association and precise localization.

\begin{table*}[t]
\centering
\begin{tabular}{@{}l|@{}rrr@{}}
\toprule
& \multicolumn{3}{c}{mvIoU} \\
Model & Slow & Medium & Fast \\
\midrule
Gemini 2.5 Pro~\cite{comanici2025gemini25} & 7.00 & 5.58 & 3.94 \\ 
GPT-5~\cite{singh2025openai} & \textbf{10.94} & \textbf{7.98} & \textbf{4.84}  \\ 
\bottomrule
\end{tabular}
\caption{Effect of velocity on LongEgoRefer.}
\label{tab:effect_of_velocity}
\end{table*}

\paragraph{Grounded-SAM2 vs SAM3}
Table~\ref{tab:replacement_with_sam3} presents experimental results on LongEgoRefer where Grounded-SAM2 was replaced with SAM3. Performance actually dropped, possibly because Grounded-SAM2's independent grounding module handles complex referring expressions more robustly.

\begin{table}[t]
\centering
\small
\scriptsize
\setlength{\tabcolsep}{2pt}
\renewcommand{\arraystretch}{0.85}
\begin{tabular}{@{}lrrr rrr}
\toprule
& \multicolumn{3}{c}{Grounded-SAM2} & \multicolumn{3}{c}{SAM3} \\
\cmidrule(lr){2-4} \cmidrule(lr){5-7}
Model & vIoU@0.1 & vIoU@0.5 & mvIoU &  vIoU@0.1 & vIoU@0.5 & mvIoU \\
\midrule
Gemini 2.5 Pro~\cite{comanici2025gemini25} & 12.01 & 4.13 & 5.49 & 5.94 & 1.26 & 2.23 \\
GPT-5~\cite{singh2025openai} & 17.48 & 6.20 & 7.89 & 9.81 & 2.13 & 3.86 \\
\bottomrule
\end{tabular}
\caption{Results of replacing Grounded-SAM2 with SAM3.}
\label{tab:replacement_with_sam3}
\end{table}

\paragraph{Cross-Benchmark Comparison}
To explicitly demonstrate the unique challenges of LongEgoRefer, we compare our results with the performance reported in the EgoMask paper~\cite{Liang_2025_ICCV_EgoMask} (Table~\ref{tab:cross_benchmark_comparison}). 
Existing models achieve reasonable performance on EgoMask's Short clips (12.15s on avg.), but their accuracy consistently decreases on the Medium (116.30s) and Long (361.32s) splits. 
On LongEgoRefer, where videos are substantially longer (45 minutes on average), performance further drops to near zero (e.g., 49.95\% to 0.67\% for Grounded-SAM2). 
This substantial drop highlights LongEgoRefer's distinct challenges, particularly extreme video duration and target sparsity.

\begin{table}[t]
\centering
\scriptsize
\setlength{\tabcolsep}{4pt}
\renewcommand{\arraystretch}{0.85}
\begin{tabular}{lrrrr}
\toprule
& \multicolumn{3}{c}{EgoMask} & LongEgoRefer \\
\cmidrule(lr){2-4} \cmidrule(lr){5-5}
Model & Short & Medium & Long & All \\
\midrule
Grounded-SAM2~\cite{ren2024grounded} & 49.95 & 25.73 & 24.80 & 0.67 \\
VideoLISA~\cite{bai2024one} & 17.85 & 6.48  & 5.15  & 0.37 \\
Sa2VA~\cite{sa2va} & 29.00 & 17.02 & 8.11  & 1.72 \\
\bottomrule
\end{tabular}
\caption{Cross-benchmark mvIoU comparison.}
\label{tab:cross_benchmark_comparison}
\end{table}

\section{Time Consumption}
\begin{figure}[t]
\centering
\includegraphics[width=\linewidth]{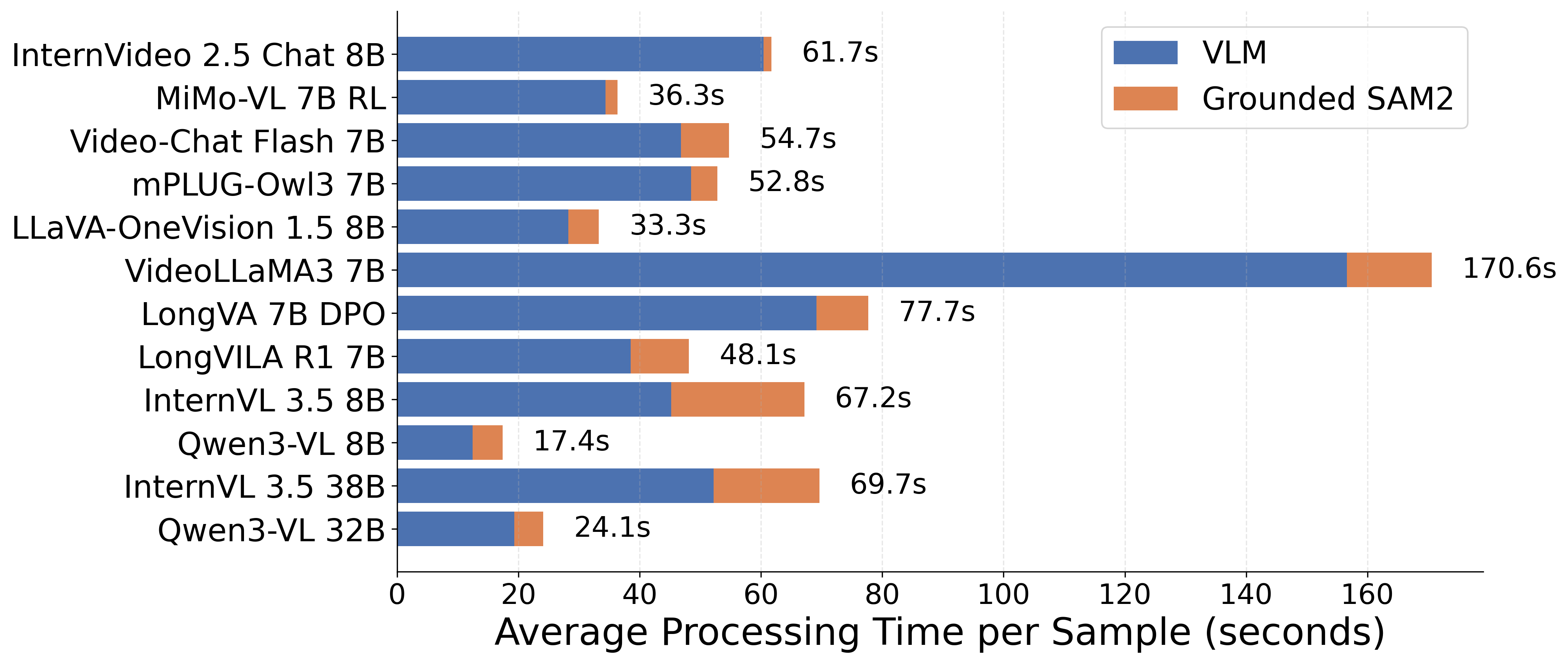}
\caption{Time consumption over the baselines.}
\label{fig:processing_time}
\end{figure}

We present the time consumption for different VLMs in Figure~\ref{fig:processing_time}.

\section{Limitations and Future Work}

Beyond the directions discussed in Section~\ref{sec:conclusion}, we note several additional limitations.
First, the extreme temporal length of our videos imposes heavy computational demands on current models. Consequently, high-resolution evaluations (e.g., higher frame rates like 5 FPS) and more exhaustive spatio-temporal inferences were constrained by these computational bottlenecks.
Second, we did not conduct a separate, controlled human-performance study, and therefore do not report a formal human-ceiling baseline. While our rigorous multi-tier quality assurance workflow was designed to ensure that the finalized queries are clear and consistently interpretable by human annotators, establishing a precise, quantitative human performance ceiling remains an important objective for future iterations of this benchmark.
Third, a potential risk is data contamination, as LongEgoRefer is built upon the widely adopted Ego4D dataset. It is possible that both the evaluated open-weight and closed-source models have been exposed to the underlying videos during their large-scale pre-training or fine-tuning phases. Although our interaction-centric referring expressions are entirely novel, this prior visual exposure remains a common limitation when evaluating modern foundation models.
Future research should therefore not only aim to improve grounding accuracy on LongEgoRefer but also explore more computationally efficient architectures and robust evaluation protocols tailored for extreme long-form video understanding.

\end{document}